# Keyword Extraction, and Aspect Classification in Sinhala, English, and Code-Mixed Content

Rizvi F.A.
*Department of Computer Science*
*Sri Lanka Institute of Information Technology*
Colombo, Sri Lanka
azmarahrizvi22@gmail.com

Navojith T.
*Department of Computer Science*
*Sri Lanka Institute of Information Technology*
Colombo, Sri Lanka
navojithamindu@gmail.com

Adhikari A.M.N.H
*Department of Computer Science*
*Sri Lanka Institute of Information Technology*
Colombo,Sri Lanka
nipunaadhikari@gmail.com

Senevirathna W.P.U.
*Department of Computer Science*
*Sri Lanka Institute of Information Technology*
Colombo, Sri Lanka
udesenevirathna@gmail.com

Dharshana Kasthurirathna
*Department of Computer Science*
*Sri Lanka Institute of Information Technology*
Colombo, Sri Lanka
dharshana.k@sliit.lk

Lakmini Abeywardhana
*Department of Computer Science*
*Sri Lanka Institute of Information Technology*
Colombo, Sri Lanka
lakmini.d@sliit.lk

*Abstract*—Brand reputation in the banking sector is maintained through insightful analysis of customer opinion on code-mixed and multilingual content. Conventional NLP models misclassify or ignore code-mixed text, when mix with low resource languages such as Sinhala-English and fail to capture domain-specific knowledge. This study introduces a hybrid NLP method to improve keyword extraction, content filtering, and aspect-based classification of banking content. Keyword extraction in English is performed with a hybrid approach comprising a fine-tuned SpaCy NER model, FinBERT-based KeyBERT embeddings, YAKE, and EmbedRank, which results in a combined accuracy of 91.2%. Code-mixed and Sinhala keywords are extracted using a fine-tuned XLM-RoBERTa model integrated with a domain-specific Sinhala financial vocabulary, and it results in an accuracy of 87.4%. To ensure data quality, irrelevant comment filtering was performed using several models, with the BERT-base-uncased model achieving 85.2% for English and XLM-RoBERTa 88.1% for Sinhala, which was better than GPT-4o, SVM, and keyword-based filtering. Aspect classification followed the same pattern, with the BERT-base-uncased model achieving 87.4% for English and XLM-RoBERTa 85.9% for Sinhala, both exceeding GPT-4 and keyword-based approaches. These findings confirm that fine-tuned transformer models outperform traditional methods in multilingual financial text analysis. The present framework offers an accurate and scalable solution for brand reputation monitoring in code-mixed and low-resource banking environments.

## I. INTRODUCTION

### A. Background

Brand reputation is one of the most valuable assets for banks. In the digital age of today, customer trust is usually developed or influenced by online conversations on social media platforms such as Twitter, YouTube, and other banking review sites. Due to the extensive usage of mobile banking and internet-based financial services, more than 70% of customer opinions are now expressed on social media and public forums [1]. To handle this feedback, banks use tools such as Meltwater[1] and Brandwatch[2] to monitor sentiment. These tools are limited to high resource languages like English and do not support low resource languages such as Sinhala and Tamil, or English–Sinhala code-mixed content, which is prevalent in Sri Lankan banking conversation. Furthermore, they rely on manual keyword input, so they are slow to pick up on emerging trends or language use. The other key problem is the poor quality of raw data. Approximately 30–40% of scraped data is not relevant and contains spam, advertisements, and system auto-generated messages [2]. It makes it hard to pay attention to real customer complaints. Conventional NLP tools are not designed to process this noisy and mixed-language data, making more sophisticated, language-aware technologies necessary.

### B. Problem Statement

Most existing brand monitoring systems cannot handle code-mixed banking text in Sinhala-English. Such models either ignore or misclassify financial terminology, leading to an absence of proper comprehension of customers' sentiments. The state-of-the-art keyword extraction techniques such as Term Frequency–Inverse Document Frequency (TF-IDF) [5] and Rapid Automatic Keyword Extraction (RAKE) [6] rely on statistical frequency and co-occurrence. The above approaches have limitations in their ability to recognize domain-specific terms, especially when the language is code-mixed or involves non-standard orthography and bank terminology [3]. Moreover, aspect classification models that are shown to be effective in English end up performing poorly when extended to Sinhala or code-mixed reviews, causing accuracy to decline. These issues reduce the effectiveness of brand monitoring tools in multilingual environments. To overcome these issues, this study presents a hybrid NLP technique that performs automatic keyword extraction, eliminates noisy data, and accurately identifies the main topics in banking customer reviews.

---

[1] https://www.meltwater.com/en
[2] https://www.brandwatch.com/

*C. Research Gap*

Despite progress in keyword extraction and aspect classification fields, several gaps still exist in handling code-mixed financial text. The main gaps are:

- **Poor performance in processing Sinhala-English code-mixed banking discussions:** Existing NLP models trained on monolingual corpora cannot capture multilingual representations, therefore misclassifying banking terminology [4].
- **Inability to differentiate between relevant and irrelevant financial content:** Current social media scraping and filtering procedures cannot differentiate actual customer discussions from spam, advertising, and bot replies [2].
- Manual Existing systems such as Meltwater[1] and Brandwatch[2] require users to enter keywords manually, which are error-prone, time-intensive, and not scalable for dynamic and evolving financial discussions.

This study aims to bridge these gaps by using domain-specific keyword extraction, data filtering methods, and transformer-based aspect classification models to improve the efficiency and accuracy of banking content processing.

*D. Objectives*

The key objectives of this research are:

1. To develop an English and Sinhala keyword extraction system using NER, FinBERT, KeyBERT, YAKE, EmbedRank, and XLM-RoBERTa.
2. To Filter irrelevant content using BERT and GPT-4.
3. To Use BERT and LLM models to categorize comments into key banking aspects, which include Customer Support, Transactions, Payments & Accounts, Loans & Credit Services, Digital Banking, and Trust & Security.

## II. LITERATURE REVIEW

Conventional keyword extraction methods are relied upon rule-based methods such as stop word elimination, stemming, and frequency-based ranking. TF-IDF (Term Frequency-Inverse Document Frequency), conceivably the most well-known of such methods, assigns weight to words based on their relative frequency in a document and a corpus [5]. Though effective for general text analysis, TF-IDF fails to determine contextual meaning and domain specificity. Another widely used approach, RAKE (Rapid Automatic Keyword Extraction), identifies keywords by detecting word co-occurrence and segmenting text into multi-word key phrases [6]. However, RAKE does not have semantic relations, which diminishes its effectiveness in code-mixed and low-resource languages such as Sinhala-English banking text [6]. Graph-based keyword extraction techniques such as TextRank extend Google's PageRank algorithm to text analysis, ranking words based on connectivity in a text corpus [7]. However, graph-based models do not address polysemy (multiple meanings for words) and code-switching, making them ineffective in multilingual finance discussions. To overcome the limitations of statistical methods, researchers introduced supervised learning models for keyword extraction. These models use annotated datasets to train classifiers such as Naïve Bayes, Support Vector Machines (SVMs), and Random Forests [8]. Although these techniques exhibit better performance, they depend on large, annotated datasets, which do not exist for low-resource languages like Sinhala. The introduction of transformer-based models, embodied in BERT (Bidirectional Encoder Representations from Transformers), has transformed the keyword extraction process. BERT-based models, such as KeyBERT and FinBERT, employ contextual embeddings to effectively capture domain-specific terminology, recording much better performance compared to conventional models [9], [11]. Research shows that FinBERT, trained in financial reports, improves keyword extraction quality by 18% over TF-IDF-based methods [10]. Another deep learning-based keyword extraction improvement is EmbedRank, in which sentence embeddings are utilized to rank key phrases concerning the document context [12], [13]. Web scraping is currently a compulsory technique for automatic online data collection, particularly in the case of customer feedback analysis.

It facilitates large-scale harvesting of data from diverse sources such as social media sites, forums, and review sites, enabling convenient sentiment analysis and trend identification [15]. Traditional scraping approaches produce high levels of irrelevant or duplicate content that need post-processing tasks to improve data relevance and verify the quality of information retrieved [14]. Even though numerous research works have been carried out on rule-based filtering methods, these kinds of methods frequently are not able to deal with irrelevant content associated with specific domains since they can't cater to subtle, context-based sophistication in text [16]. To prevent this drawback, recent research emphasis has been on utilizing deep learning models for automatic filtering and categorization of content. Research demonstrates that BERT fine-tuning on domain-specific corpora significantly improves classification accuracy, particularly for technical domains such as banking and finance [17]. However, while BERT-based models effectively capture semantic meaning, they are often hindered by real-world user-generated text, which is comprised of informal text, abbreviations, and contextual ambiguities. As such, alternative evaluation strategies, such as model ensembling and transfer learning, are being explored to boost classification robustness [18], [19]. The advent of large language models (LLMs) like GPT [21] has added more interest in classification methods using APIs. These models, developed with massive datasets comprising multiple languages and intended domains, provide better contextual comprehension and flexibility. In research, it has already been seen that LLM APIs perform better than conventional deep learning models in various text classification tasks, specifically where the setting includes informal or highly contextual content [20]. Research furthermore shows that combining pre-trained transformer-based models and LLM APIs can guarantee further generalization, proposing a hybrid approach to filtering out unwanted content and improving aspect classification in complex datasets [20]. Despite the above advances, an investigation into the relative effectiveness of LLMs and fine-tuned transformer models for bank-related dialogue, particularly in low-resource languages such as Sinhala, remains a gap in current research [22]. This paper fills these gaps by creating a hybrid keyword extraction system that consists of a fine-tuned SpaCy NER model, FinBERT-enhanced KeyBERT, YAKE, and EmbedRank for English, and a fine-tuned XLM-RoBERTa model with a Sinhala financial dictionary for Sinhala and code-mixed data.

Similarly, by incorporating BERT-base-uncased content filtering and LLM API tests, that is, GPT, to identify the best strategy for the removal of irrelevant comments and aspect classification. Through a comparative performance comparison between English and Sinhala banking forums, this study seeks to bridge the gap between conventional deep learning methodologies and contemporary LLM-based solutions.

### III. METHODOLOGY

This research presents a hybrid NLP approach for keyword extraction and aspect classification in Sinhala-English banking content. The system integrates machine learning and rule-based methods for efficient domain-specific keyword tagging and comment categorization.

Fig.1 illustrates the end-to-end system workflow — from URL input and keyword extraction to aspect categorization.

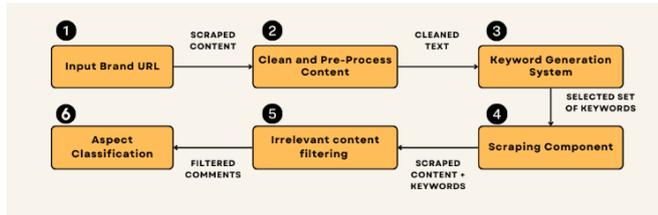

Fig. 1. Overall System Flow Chart

The process consists of three main phases:

1) **Keyword Extraction:** Utilized SpaCy NER, YAKE, FinBERT-enhanced KeyBERT, and EmbedRank for English. XLM-RoBERTa with Sinhala financial dictionary for Sinhala English code-mixed text.

2) **Scraping & Filtering:** Scraped YouTube comments via API. Removed irrelevant text utilizing BERT and GPT-4o.

3) **Aspect Categorization:** BERT and GPT-4o divided comments into six banking aspects, including Customer Support, Transactions, Payments & Accounts, Loans & Credit Services, Digital Banking, and Trust & Security.

#### A. Data Gathering and Pre-Processing

The dataset for the keyword generation system was constructed by scraping financial content from major Sri Lankan banking websites (e.g., HNB[3], BOC[4]). A total of 6000 records were collected, with 10 000 manually labeled Named Entity Recognition (NER) tags and keywords for Sinhala and English both. To ensure high accuracy, the extracted keywords and NER entities were manually adjusted and validated. This involved Correcting wrongly labeled entities, normalizing financial terminology among banks and adding missing domain-specific keywords. Furthermore, the class imbalance was handled using oversampling and undersampling techniques. This balanced dataset improved the performance of both NER models and keyword extraction tools.

Table I shows a sample example of the manually annotated dataset for aspect-based sentiment analysis for the banking domain. Every record contains a user-given comment and its respective annotated aspect. The dataset consists of 8,000 English comments and 5,000 Sinhala-English code-mixing comments gathered using keyword-based web scraping from YouTube. The comments are categorized into one of five pre-established banking aspects: Trust and Security, Digital Banking, Customer Support, Loans and Credit Services, Transactions, Trust & Security, and Payments & Accounts. The categories were validated by manual inspection to provide higher accuracy and relevance.

TABLE I
SAMPLE OF MANULLY ANNOTATED DATASET

| Comment | Aspect |
|---|---|
| The mobile banking app crashes every time I try to transfer funds | Digital Banking |
| I applied for a personal loan, but the approval process took over a month. | Loans and Credit Services |
| කස්ටමර් සපෝර්ට් එකට phone එක pickup කරන්න godak time ගියා | Customer Support |
| Payment eka process una kiyala thiyenawa, habai money eka debit wela na | Transactions |

#### B. Keyword Extraction and Generation

Accurate keyword extraction is crucial in multilingual bank customer feedback analysis. The present study adopted a two-strategy approach, handling English and Sinhala/code-mixed content separately because of the differing linguistic and resource demands. Statistical, transformer-based, and NER-driven approaches were integrated to extract domain-specific, high-quality financial keywords.

*1) English Keyword Extraction*

For English text, a multi-step hybrid solution was implemented. As the first step, a Named Entity Recognition (NER) model was trained with SpaCy's en_core_web_sm pipeline, augmented with 3,000 manually tagged banking-related sentences. This model was fine-tuned to pull out structured financial entities from discussions and reviews by customers. It successfully recognized bank names such as *HNB*, *DFCC*, and *Seylan*; loan types including *Mortgage Loan* and *Business Loan*; account types like *Savings Account* and *Fixed Deposit*; and regulatory or policy-related terms such as *Financial Policy* and *Interest Rate*. In addition to NER, three keyword extraction techniques were included to enhance coverage and accuracy of context. YAKE ranked keywords using term frequency, co-occurrence, and position, effectively identifying terms like *"savings account"* and *"interest rates"* from short reviews. FinBERT enhanced KeyBERT combined financial-domain embeddings with contextual phrase extraction, allowing it to capture semantically rich but low-frequency terms such as *"loan restructuring"*. EmbedRank, leveraging msmarco-distilbert-base-v3, computed cosine similarity between phrases and the full document, proving effective for long-form content. For instance, in *"Customer support delayed my loan approval,"* it extracted both *"customer support"* and *"loan approval"* with high precision.

---
[3] https://www.hnb.lk/
[4] https://www.boc.lk/

To rank the most relevant keywords highest, a weighted scoring system was used, which depended on the number of extraction methods that identified a particular keyword as well as the relative accuracy of each method. The composite score for every keyword K was computed using the following formula:

$$Score(K) = 2 \cdot YAKE(K) + 3 \cdot KeyBERT(K) + 4 \cdot EmbedRank(K) \quad (1)$$

These weights were assigned based on the observed accuracy of each method, with EmbedRank showing the highest contextual precision, followed by KeyBERT and YAKE. To validate keyword relevance and ensure domain alignment, top-scoring keywords were filtered through two layers: matching against fine-tuned NER outputs to confirm semantic significance, and cross-checking with a curated English financial vocabulary[5]. Keywords present in the vocabulary were retained and boosted, while unrelated terms were discarded.

*2) Sinhala Keyword Extraction*

Sinhala pipeline was specially designed to address the challenges of low-resource language processing, morphological richness, and code-mixed content prevalence in user-generated financial discussions. A xlm-roberta-large model was fine-tuned on a dataset of 3000 Sinhala sentences on banking and 5,200 manually labeled financial entities to extract meaningful financial terms from Sinhala and Sinhala-English code-mixed text. This model demonstrated proficiency in correctly recognizing complex financial frameworks, such as Sinhala banking institutions (e.g., සම්පත් බැංකුව, මහජන බැංකුව), loan types (e.g., නිවාස ණය, අධ්‍යාපන ණය), account types (e.g., ළමා ඉතුරුම් ගිණුම, අරමුදල් ගිණුම), and regulatory terms (e.g., මූල්‍ය ප්‍රතිපත්ති, ලීසිං). Unlike in English, conventional statistical methods like TF-IDF and even YAKE were less effective for Sinhala because of extended word forms, compound structures, and informal spelling variations prevalent in user-generated text. Consequently, keyword extraction for Sinhala depended mostly on contextual transformer embeddings produced by the XLM-R model itself. The output of the model consisted of named entities as well as contextually relevant keywords. To validate and refine the extracted keywords, each term was cross-checked against a curated Sinhala financial vocabulary[6]. This procedure guaranteed the preservation of low-frequency yet important terms that are specific to the domain, e.g., "නිවාස ණය" (housing loan) and "අධ්‍යාපන ණය" (educational loan), which would be disposed to removal by methods relying on frequency only. In the case of a keyword matching a term in the vocabulary, a boosting function was applied, which caused its final score to be doubled to account for its importance in the financial domain. The modified scoring approach facilitated the effective and efficient retrieval of Sinhala and code-mixed keywords that are crucial for subsequent aspect and sentiment analysis.

*C. Web Scraping, Content Filtering and Aspect Classification*

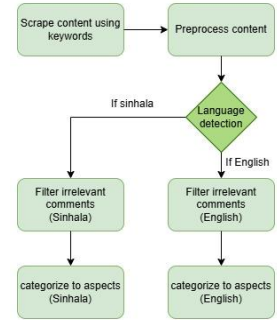

Fig. 2.Web Scraping, Content Filtering and Aspect Classification Flow Chart

The system utilizes YouTube API to web scrape comments of users on banking-related videos, for instance reviews of financial products, customer testimonials, and discussions on financial news. Boolean queries developed from banking keywords are used to web scrape the comments. Removing duplicates, cleaning symbols, and correcting formatting are preprocessing tasks that are undertaken to structure the dataset and prepare it for later processing stages. Removal of spam, ads, and off-topic content is an essential task to eliminate unnecessary comments and retain only valuable banking comments for further processing. In English and Sinhala comment processing, different approaches are tested to find the best approach for each language.

*1) English Irrelevant Comment Removal*

Irrelevant English comments are eliminated by using a combination of keyword extraction and machine learning models. A trained SVM and fine-tuned BERT-base-uncased model define relevance in terms of banking context. Furthermore, GPT API is used for contextual assessment. Performance is measured using accuracy, recall, and F1-score to select the optimal approach.

*2) Sinhala Irrelevant Comment Removal*

Sinhala comments are first filtered through keyword-based banking terminology and a fine-tuned XLM-RoBERTa model trained on Sinhala banking text. Context filtering is conducted using the GPT API. Methods are evaluated on accuracy, recall, and F1-score to select the optimum approach.

Comments of interest are then categorized into six pre-established banking aspects, with several models tested to identify the most effective classifier.

*3) English Aspect Classification*

English comment aspect classification starts with a keyword-based method, where certain banking terms are mapped to their respective aspects. This process is followed by using a Random Forest classifier that assesses the features extracted from the comments to sort them into the respective banking aspects. Furthermore, a fine-tuned BERT-base-uncased model is used to enhance the precision of the classification process. Finally, the GPT API is also utilized in aspect classification through its advanced language understanding to categorize comments based on context. Each method's performance is measured based on accuracy and consistency, with the top-performing model being utilized for final output.

---

[5]https://drive.google.com/file/d/1ZJg9ruX7s49IP-OsaoxvXzYuXJr67BRC/view?usp=sharing
[6]https://drive.google.com/file/d/1__lw60yNZ4li2x_zk3Eg8HpkKGH0Lkdd/view?usp=sharing

*4) Sinhala Aspect Classification*

For Sinhala comments, aspect classification is started by keyword-based classification using a Sinhala lexicon appropriate to a specific domain for mapping comments against suitable aspects. The fine-tuned XLM-RoBERTa-base model is then applied to enhance classification precision with proper regard for the nuanced nature of Sinhala banking discussions. Furthermore, the GPT API is used for classifying the remaining content, making use of its ability to recognize advanced linguistic patterns. As in the English aspect classification, evaluation metrics like accuracy and F1-score are employed to compare the performance of each approach. Based on these measures, the best-performing model in the Sinhala aspect classification is chosen.evaluations.

IV. RESULTS AND DISCUSSION

*A. Keyword Extraction Accuracy*

As indicated in Table II, the top-performing keyword extraction model was the hybrid keyword extraction model with an accuracy of 91.2% and F1-score of 90.1%, topping all the baseline and single models implemented in the study. Despite FinBERT-augmented KeyBERT and EmbedRank achieving high single accuracy of 83.5% and 85.2% respectively, they failed to beat the overall precision of the hybrid model, which successfully integrates several extraction indicators. Traditional methods like TF-IDF and RAKE underperformed, showing their limited capacity to capture domain-specific context.

TABLE II
ENGLISH KEYWORD EXTRACTION RESULTS

| Methodology | Accuracy (%) | F1-Score (%) |
|---|---|---|
| TF-IDF | 61.2 | 59.3 |
| RAKE | 64.8 | 61.7 |
| YAKE | 72.1 | 70.3 |
| KeyBERT (FinBERT) | 83.5 | 81.9 |
| EmbedRank | 85.2 | 83.6 |
| **Hybrid Method** | **91.2** | **90.1** |

Table III illustrates that the performance of Sinhala Keyword Extraction demonstrates that the fusion of a fine-tuned XLM-RoBERTa model with a specialized financial vocabulary enrichment mechanism achieved the best accuracy of 87.4% and an F1-score of 86.1%. This result outperformed traditional statistical approaches like TF-IDF (51.4%) and RAKE (55.3%), which were confronted with morphological discrepancies and code-mixed grammar. The vocabulary validation component was essential in enhancing accuracy by fostering infrequent yet valuable words such as "මුදල් ආපසු ගැනීම" (fund withdrawal) and "මූල්‍ය අර්බුදය" (financial crisis).

TABLE III
SINHALA KEYWORD EXTRACTION RESULTS

| Methodology | Accuracy (%) | F1-Score (%) |
|---|---|---|
| TF-IDF | 51.4 | 48.9 |
| RAKE | 55.3 | 52.1 |
| Modified YAKE for Sinhala | 66.4 | 63.2 |
| XLM-RoBERTa | 83.1 | 81.7 |
| **Fine-Tuned XLM-R + Vocabulary Boosting** | **87.4** | **86.1** |

*B. Irrelevant Comment Removal*

As per Table IV, some models were tested for the removal of irrelevant comments in English banking feedback, such as Support Vector Machine (SVM), BERT-base-uncased, GPT-4 API, and a rule-based keyword matching model. The models were tested on accuracy, recall, and F1-score. Out of these, the BERT-base-uncased model performed better than the others, with the highest accuracy and F1-score, proving to be the best in detecting irrelevant content.

TABLE IV
ENGLISH IRRELEVANT CONTENT REMOVAL RESULTS

| Model | Accuracy (%) | Precision (%) | Recall (%) | F1-score (%) |
|---|---|---|---|---|
| BERT-base-uncased | 85.2 | 83.2 | 84.5 | 83.8 |
| GPT-4 API | 79.1 | - | - | - |
| SVM | 76.2 | 75.4 | 77.5 | 76.4 |
| Keyword-based | 72.5 | 71.3 | 73.2 | 72.6 |

For a comparison, results for Sinhala and code-mixed feedback are given in Table V, where the XLM-RoBERTa model proved to be the best performing. It achieved the best accuracy and F1-score and surpassed SVM, GPT-4 API, and rule-based approaches. This is mainly since it possesses cross-lingual contextual embeddings and can effectively handle morphologically rich and code-mixed syntax.

TABLE V
SINHALA IRRELEVANT CONTENT REMOVAL RESULTS

| Model | Accuracy (%) | Precision (%) | Recall (%) | F1-score (%) |
|---|---|---|---|---|
| XLM-RoBERTa Model | 85.9 | 84.3 | 87.1 | 85.7 |
| GPT-4 API | 76.8 | - | - | - |
| Keyword-based | 74.6 | 73.0 | 76.0 | 74.4 |

*C. Aspect Classification Accuracy*

As indicated in Table VI, the BERT-base-uncased model performed better in English aspect classification with high accuracy and F1-score.

TABLE VI
ENGLSH ASPECT CLASSIFICATION RESULTS

| Model | Accuracy (%) | Precision (%) | Recall (%) | F1-score (%) |
|---|---|---|---|---|
| BERT-base-uncased | 87.4 | 85.2 | 86.0 | 85.6 |
| GPT-4 API | 79.1 | - | - | - |
| Random Forest | 77.2 | 75.3 | 76.9 | 75.0 |
| Keyword-based | 73.4 | 72.3 | 74.8 | 72.5 |

As shown in Table VII, XLM-RoBERTa-base performed best in aspect classification for both Sinhala and code-mixed data with the highest accuracy and F1-score compared to all other models.

TABLE VII
SINHALA ASPECT CLASSIFICATION RESULTS

| Model | Accuracy (%) | Precision (%) | Recall (%) | F1-score (%) |
|---|---|---|---|---|
| XLM-RoBERTa Model | 88.1 | 86.2 | 87.5 | 85.8 |
| GPT-4 API | 77.1 | - | - | - |
| Keyword-based | 74.5 | 73.3 | 71.0 | 73.6 |

BERT-base-uncased was used for English due to its optimization for monolingual English tasks, while XLM-RoBERTa was chosen for Sinhala and code-mixed data because of its multilingual capabilities, including Sinhala script support. Each model was selected based on language compatibility to ensure optimal accuracy.

*C. Limitations*

Despite strong performance, the system faces several limitations. The lack of high-quality annotated Sinhala financial datasets restricts deeper model training. Code-mixed Sinhala-English content remains challenging due to its informal structure and limited labeled data. Transformer models, while accurate, are computationally intensive and less efficient with lengthy texts. Additionally, web scraping introduces noise, requiring improved filtering techniques.

V. CONCLUSION AND FUTURE WORK

This study introduced a robust multilingual brand reputation monitoring system tailored for the banking sector, focusing on automated keyword extraction, irrelevant content filtering, and aspect classification. The keyword generation component effectively combined statistical techniques, domain-specific contextual models (FinBERT-enhanced KeyBERT), and fine-tuned NER, further refined through financial vocabulary validation and a custom scoring mechanism. On classification tasks, BERT-base-uncased and XLM-RoBERTa yielded superior results on English and Sinhala datasets, respectively, than GPT-4o API and baseline models in accuracy and reliability. Future improvements will involve the integration of a sentiment classification module for Sinhala-English financial discussions to facilitate aspect-based sentiment analysis end-to-end. Explainability measures will also be integrated using SHAP and LIME methods to enhance model transparency and stakeholder trust. The system will also be expanded further to support Tamil and other low-resource languages, thus ensuring more regional acceptability. Moreover, there are plans to improve social media scraping capabilities to include social media platforms like Facebook and Twitter to enable the harvesting of a broader and more diverse range of multilingual financial feedback in real time.